Noname manuscript No.
In 2012

# Robust Subspace Outlier Detection in High Dimensional Space

Zhana

***Abstract*—Rare data in a large-scale database are called outliers that reveal significant information in the real world. The subspace-based outlier detection is regarded as a feasible approach in very high dimensional space. However, the outliers found in subspaces are only part of the true outliers in high dimensional space, indeed. The outliers hidden in normal-clustered points are sometimes neglected in the projected dimensional subspace. In this paper, we propose a robust subspace method for detecting such inner outliers in a given dataset, which uses two dimensional-projections: detecting outliers in subspaces with local density ratio in the first projected dimensions; finding outliers by comparing neighbor's positions in the second projected dimensions. Each point's weight is calculated by summing up all related values got in the two steps projected dimensions, and then the points scoring the largest weight values are taken as outliers. By taking a series of experiments with the number of dimensions from 10 to 10000, the results show that our proposed method achieves high precision in the case of extremely high dimensional space, and works well in low dimensional space.**

***Keywords-Outlier* detection; High dimensional subspace; Dimension projection; k-NS;**

## I. INTRODUCTION

Finding rare and valuable data is always a significant issue in data mining field. These worthy data are called anomaly data that are different from the rest of the normal data based on some measures. They are also called outliers that are located far in distance from others. Outlier detection has many practical applications in different domains, such as medicine development, fraud detection, sports statistics analysis, public health management, and so on. According to different perspectives, many definitions about outliers are proposed. The widely accepted definition is Hawkins': an outlier is an observation that deviates so much from other observations as to arouse suspicion that it was generated by a different mechanism[7]. This definition not only describes the difference of data from observation but also points out the essential difference of data in mechanism; even though some synthetic data are generated according to this concept in order to verify their outliers' detection methods.

Although outlier detection itself does not have a special requirement for high dimensional space, large-scale data are more practicable in the real world. There are two issues for outlier detection in high dimensional space: the first one is to overcome the complexity in high dimensional space, and the other is to meet the requirement for real applications with the tremendous growth of high dimensional data. In low dimensional space, outliers can be considered as far points from the normal points based on the distance. However, in high dimensional space, the distance no longer meets the exact description between outliers and normal data. In this case, detecting outliers falls into two categories, distance-based and subspace-based methods. The first one uses robust distance or density in high dimensional space, i.e. LOF[1], Hilout[8], LOCI[3], GridLOF[14], ABOD[4], etc. These methods are suitable for the outlier detection in not high dimensional space. However, in very high dimensional space, they perform poor because of "curse of dimensions". The other one that subspace based detection is an optimum method to find outliers in high dimensional space. It is based on the assumption that the outliers in all low projected dimensional subspaces are taken as real outliers in high dimensional space. This solution includes Aggarwal's Fraction[2], GLS-SOD[16], CURIO[15], SPOT[22], Grid-Clustering[23], etc. Since outliers are easily found in low projected dimensions using some optimized search algorithms to find suitable cell-grids that are divisions of subspace, it is widely used for outlier detection in high dimensional space. Recent advance in geo-spatial, bio-informatics, genetics and particle physics also require more robust subspace detection methods in growing high dimensional data. However, one key issue is still uncertain:

- *Is that truth that the outliers detected in subspaces are all outliers in high dimensional space?*

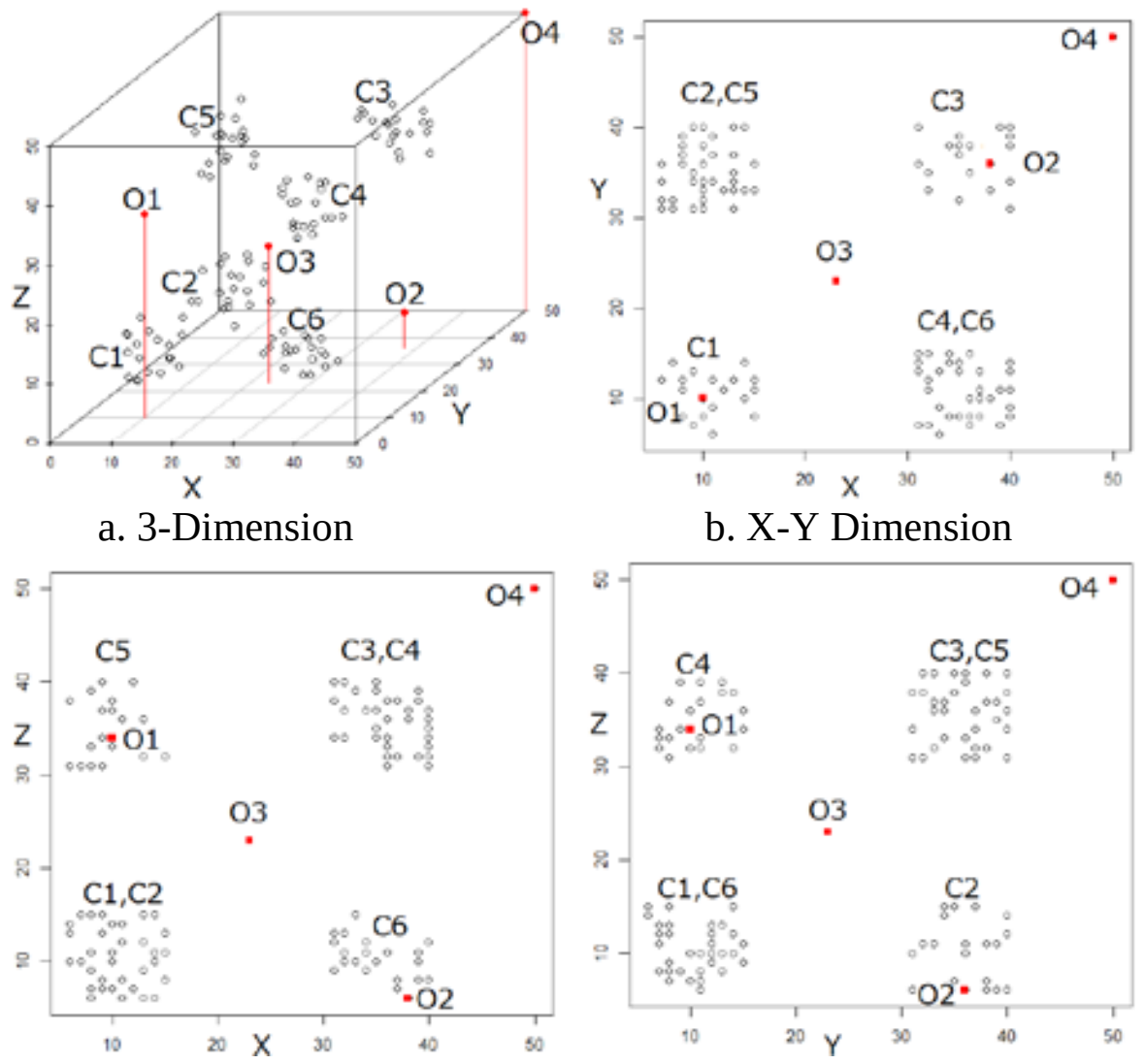

a. 3-Dimension b. X-Y Dimension

c. X-Z Dimension d. Y-Z Dimension

Figure 1. Sample data plotted in three-dimensional space and in two-dimensional spaces. Four red outliers separated in (a) are observed. But in (b), (c) and (d), only two red outliers are observed. Other two outliers are hidden in the normal clusters.

In fact, subspace-based detection methods can find some outliers different from normal points in projected dimensional space, but they ignore the outliers hidden inside the region of normal data. These inner outliers are still different from normal data in high dimensional space. We show a simple example to prove the difference between these two types of outliers separately in three-dimensional space and projected two-dimensional subspaces, as shown in Fig. 1. Total 124 points are distributed in a three-dimensional space, including 120 normal points in six clusters and 4 outliers in red color. In (a), four outliers can be found differently because they do not belong to any normal clusters. The outliers $O_3$ and $O_4$ are detected different in any of the projected dimensional spaces, while the inner outliers $O_1$ and $O_2$ are hidden inside the clusters in the projected dimensional space. Therefore, detecting $O_1$ and $O_2$ fails. All subspace-based methods fail to detect these inner outliers, as shown in (b), (c) and (d). From the above, how to find all outliers with subspace-based method is still an issue to be considered.

In this paper, we try to solve this issue by utilizing the two dimensional-projections and propose a robust subspace detection method called k-NS(k-Nearest Sections). It calculates *ldr*(local density ratio) in the first projected dimensional subspace and the nearest neighbors' *ldr* in the second projected dimensional subspace. Then, each point's weight is summed statistically. The outliers are those scoring the largest weights.

The main features and contributions of this paper are summarized as follows:

- We apply two dimensional-projections to calculate the weight values in all projected dimensions. For each point, we supply the ($m + m \times (m-1)$) weight values in order to compare it with others extensively.
- Our proposed method employs k-NS(k-Nearest Sections) based on the k-NN (k Nearest Neighbor) concept for the local density calculation in the second projected dimensional space. The inner outliers are detected successfully by evaluating the neighbor's *ldr* after projecting them into other dimensions.
- We execute a series of experiments with the range of dimensions from 10 to 10000 to evaluate our proposed algorithm. The experiment results show that our proposed algorithm has advantages over other algorithms with stability and precision in high dimensional dataset.
- We also consider the difference between the outliers and noisy data. The outliers are obviously different in high dimensional space with noisy data while they are mixed together in low dimensional space.

This paper is organized as follows. In section 2, we give a brief overview of related works on high dimensional outlier detection. In section 3, we introduce our concept and our approach, and we describe our algorithm. In section 4, we evaluate the proposed method by experiments of different dimensional datasets, artificially generated and real datasets. At last, we conclude our findings in section 5.

## II. RELATED WORKS

As an important part of the data mining, outlier detection has been developed for more than ten years, and many study results have been achieved in large scale database. We categorize them into the following five groups.

**Distance and Density Based Outlier Detection**: the distance based outlier detection is a conventional method because it comes from the original outlier definition, i.e. outliers are those points that are far from other points based on distance measures, e.g. by Hilout[8]. This algorithm detects point with its k-nearest neighbors by distance and uses space-filling curve to map high dimensional space. The most well known LOF[1] uses k-NN and density based algorithm, which detects outliers locally by their k-nearest distance neighbor points and measures them by lrd (local reachability density) and lof(Local Outlier Factor). This algorithm runs smoothly in low dimensional space and is still effective in relative high dimensional space. LOCI[3] is an improved algorithm based on LOF, which is more sensitive to local distance than LOF. However, LOCI does not perform well as LOF in high dimensional space.

**Subspace Clustering Based Outlier Detection**: since it is difficult to find outliers in high dimensional space, they try to find these points behaving abnormally in low dimensional space. Subspace clustering is a feasible method for outlier detection in high dimensional space. This approach assumes that outliers are always deviated from others in low dimensional space if they are different in high dimensional space. Aggarwal[2] uses the equi-depth ranges in each dimension with expected fraction and deviation of points in k-dimensional cube D given by $N \times f^k$ and $\sqrt{N \times f^k \times (1 - f^k)}$. This method detects outliers by calculating the sparse coefficient S(D) of the cube D.

**Outlier Detection with Dimension Deduction**: another method is dimension deduction from high dimensional space to low dimensional space, such as SOM (Self-Organizing Map)[18,19], mapping several dimensions to two dimensions, and then detecting the outliers in two dimensional space. FindOut[11] detects outliers by removing the clusters and deducts dimensions with wavelet transform on multidimensional data. However, this method may cause information loss when the dimension is reduced. The result is not as robust as expected, and it is seldom applied to outlier detection.

**Information-theory based Outlier Detection**: in subspace, the distribution of points in each dimension can be coded for data compression. Hence, the high dimensional issue is changed to the information statistic issue in each dimension. Christian Bohm has proposed CoCo[9] method with MDL(Minimum Description Length) for outlier detection, and he also applies this method to the clustering issue, e.g. Robust Information-theory Clustering[5,12].

**Other Outlier Detection Methods**: besides above four groups, some detection measurements are also distinctive and useful. One notable approach is called ABOD (Angle-Based Outlier Detection)[4]. It is based on the concept of angle with vector product and scalar product. The outliers usually have the smaller angles than normal points.

| Point NO | X | Y |
|---|---|---|
| pt1 | 8 | 6 |
| pt2 | 10 | 7 |
| pt3 | 6 | 8 |
| pt4 | 6 | 6 |
| pt5 | 7 | 9 |
| pt6 | 10 | 8 |
| pt7 | 8 | 7 |
| pt8 | 9 | 8 |
| pt9 | 8 | 9 |
| pt10 | 8 | 8 |
| pt11 | 15 | 15 |
| pt12 | 23 | 24 |
| pt13 | 23 | 21 |
| pt14 | 22 | 23 |
| pt15 | 20 | 21 |
| pt16 | 21 | 24 |
| pt17 | 21 | 20 |
| pt18 | 22 | 21 |
| pt19 | 22 | 20 |
| pt20 | 22 | 22 |
| pt21 | 23 | 21 |
| pt22 | 5 | 25 |
| pt23 | 20 | 10 |

(a)

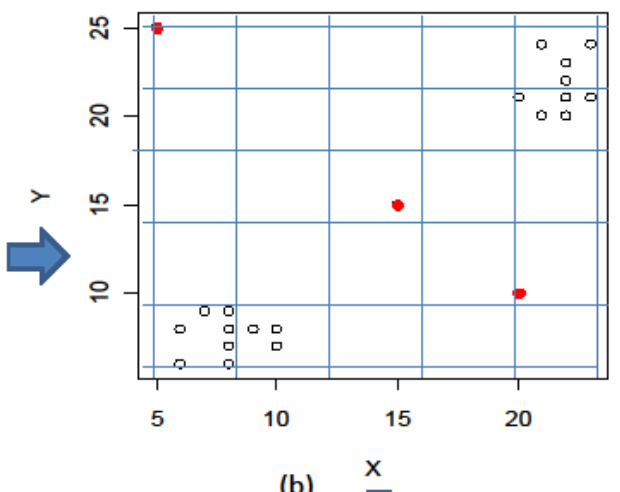

(b)

| | section1 | section2 | section3 | section4 | section5 |
|---|---|---|---|---|---|
| x | 8 | 3 | 1 | 0 | 11 |
| y | 10 | 1 | 1 | 6 | 5 |

(d)

| Point NO | X | Y |
|---|---|---|
| pt1 | 1 | 1 |
| pt2 | 2 | 1 |
| pt3 | 1 | 1 |
| pt4 | 1 | 1 |
| pt5 | 1 | 1 |
| pt6 | 2 | 1 |
| pt7 | 1 | 1 |
| pt8 | 2 | 1 |
| pt9 | 1 | 1 |
| pt10 | 1 | 1 |
| pt11 | 3 | 3 |
| pt12 | 5 | 5 |
| pt13 | 5 | 4 |
| pt14 | 5 | 5 |
| pt15 | 5 | 4 |
| pt16 | 5 | 5 |
| pt17 | 5 | 4 |
| pt18 | 5 | 4 |
| pt19 | 5 | 4 |
| pt20 | 5 | 5 |
| pt21 | 5 | 4 |
| pt22 | 1 | 5 |
| pt23 | 5 | 2 |

(c)

Figure 2. Section Space Division and Dimension Projection

The above methods have reduced the "high dimensional cures" to some extent, and they get the correct results in some special cases. However, the problem still exists and affects the point's detection accuracy. Christian Bohm's information-theory based method is similar to the subspace clustering methods and suffers the same with subspace-based outlier detection methods.

In summary, seeking a general approach or improving the existed subspace-based methods to detect outliers in high dimensional space is still a key issue needs to be solved.

## III. PROPOSED METHOD

It is known from last section that not all outliers can be found in projected dimensional subspace. The outliers failing to be detected in subspace are called inner outliers. The inner outliers are mixed in normal clusters in projected dimensional subspaces, but they are detected anomaly in high dimensional space. From another point of view, the inner outliers belong to several normal clusters in different subspaces, but they do not belong to any cluster as a whole. In this paper, the key mission is to find such inner outliers in high dimensional space.

### *A. General Idea*

Learning from the subspace detection methods, we know that high dimensional issue can be transformed into the statistical issue by loop detection in all projected dimensional subspaces. Moreover, the points' distribution is independent in different dimensions. By observing these points and learning from existing outlier definitions, we have found that the outliers are placed in a cluster of normal points in a certain dimension and deviated in other dimensions. Otherwise, outliers are clustered with different normal points in different dimensions while normal points are always clustered together. Therefore, our proposed method needs to solve the two sub-issues: how to find outliers effectively in all projected-dimensional subspaces; and how to detect the deviation of points of the same region in one dimension when these points are projected to other dimensions.

Our proposal can be divided into four steps. First, we divide the entire range of data into many small regions in each dimension. Here, we call the small region a section. Based on the section division, we construct the new data structure called section space. Second, we calculate the sparsity of point in each section in each dimension by computing the *ldr* against average value in that dimension. Third, we calculate the scattering of the same section points by *ldr* after projecting them from original dimension to other dimensions. Last, we sum up all the results as weight for each point, and then compare all the points with the score. The outliers are the points scoring the largest values of weight.

### *B. Section Data Structure*

Our proposed method is based on the section data structure. The mechanism on how to compose this section structure and transform the Euclidean data space into our proposed section space is introduced in below.

We divide the space into the same number of equi-width sections in each dimension, so the space just looks like a cell-grid. The conventional data space information is composed of points and dimensions while our proposed data structure represents the data distribution by point, dimension and section. This structure has two advantages. First, the point in which section is easily found out in all dimensions. Therefore, we can use all related section calculated result to denote the point's weight value. Second, it is easy for calculating the distribution change by checking the points' section position while projecting them to different dimensions.

The data structure of *PointInfo*(point information) and *SectionInfo*(section information) cited in our proposal are shown as follows:

***PointInfo****[Dimension ID, Point ID]: section ID* of the *point*
***SectionInfo****[Dimension ID, Section ID]: #points in the section*

The *PointInfo* records each point's section position in different dimensions. The *SectionInfo* records the number of points of each section in different dimensions. The main calculation of sparsity of points and dimension projections are processed based on these two data structures. The transforming process from the original data space to the proposed section-based space is explained using an example of two-dimensional dataset, as shown in Fig. 2. The dataset includes 23 points in two-dimensional space, as shown in Fig. 2(a). The original data distribute in the data space based on Euclidean distance is shown in Fig. 2(b). In our proposed section-based structure, we construct the *PointInfo* structure as in Fig. 2(c) and *SectionInfo* structure in as in Fig. 2(d). The range of each dimension is divided into five sections in this example. The section division is shown in Fig. 2(b) with blue lines.

The data range of each dimension may be different. If we set the same data range for every dimension covering the maximum of a certain dimension, it would produce too many empty sections in some dimensions. The empty sections producing meaningless values 0 would affect the result markedly in the following calculations. Therefore, we set the minimum data range in each dimension covering the only area where points exist. In order to avoid two end-sections having larger density than that of other sections, we extend the border by enlarging the original range by 0.1%. Taking the data in Fig. 2 as an example to explain how to generate the data range in each dimension, the original data range in *x* dimension is (5, 23), and the length is 18. The new extended-range is (4.991, 23.009) by enlarging the length by 0.1%. Therefore, the new length is 18.018. The original data range in *y* dimension is (6, 25), and the length is 19. The new data range is (5.9905, 25.0095), and the new length is 19.019. The length of a section is 3.6036 in x dimension and 3.8038 in *y* dimension.

## C. *Definitions*

To our proposal, some definitions of notations are given in Table 1:

Table 1. Definition of Notations

| Symbol | Definition |
|---|---|
| ***P*** (point) | The information of point. $\boldsymbol{p_j}$ refers to the j[th] point of all points. $\boldsymbol{p_{i,j}}$ refers to the j[th] point in i[th] dimension. |
| ***Section*** (section) | The range of data in each dimension is divided into the same number of equi-width parts, which are called sections. |
| ***scn*** (number of section) | The number of sections for each dimension. It is decided by the number of total points and the average section density. *scn* is defined equally in each dimension. |
| ***d*** (section density) | The number of points in one section is called section density, *d* for short. |
| ***dists*** (section distance) | The section distance used for evaluating the section difference among points in all projected dimensions, as defined in (1). |
| ***ldr*** | Local density ratio, after introducing the Section, it is replaced by *sdr* |
| ***sdr*** | Section density ratio, the calculation is defined in (4) and (5) |
| ***SI*** (statistic information) | The statistic information of each point composed of all weights, as defined in (6). |

The section density *d* with different subscriptions presents specific meanings in following cases.

- Case 1: in a section, all points of this section in a dimension have the same section density, and $d_{i,j}$ means section density value for the points in the $j^{th}$ section in the $i^{th}$ dimension.
- Case 2: the section density is used to compare it with the average density in this dimension. So the low section-density means a low ratio against the average section density in a dimension. $\overline{d_i}$ means the average section density in the $i^{th}$ dimension.
- Case 3: if the section density of a point is needed, the expression will include the point. $d_i(p)$ means the section density value of point p in the $i^{th}$ dimension.

In section-based subspace, the section denotes the point's local area. The local density is replaced by *d*. Then the *ldr* is replaced by *sdr*(section density ratio).

The process of two dimensional-projections is introduced in our proposal. Projecting points to each one-dimensional subspace is the first projection. All points are checked in all the projected dimensional subspaces. After that, the points in projected dimensions still need to be checked between different subspaces in order to detect inner outliers. Therefore, the points are projected again from the first projected dimension to other dimensions, and then compare their distribution changes with each other. It is called the second dimension projection. The whole procedure projects the points twice: from high dimension to one dimension and from one dimension to other dimensions.

## D. *k-Nearest Sections*

In this section, we describe the detection methods in two steps. In the first step, the *sdr* is employed to evaluate the sparsity of points in first projection dimensions. In the second step, which is the key part of this proposal, the scattering of points after their second projections to other projected dimensions is calculated based on k-NS(k-Nearest Sections). At last, we summarize these results of two steps statistically.

Before introducing the concept of k-NS, the *dists* needs to be clarified in advance.

**DEFINITION 1** (*dists* of points)
Let point $p,q \in Section$ . $p_i, q_i$ are in the $i^{th}$ dimension. When $p_i, q_i$ are projected from the dimension *i* to *j*, the section distance between them corresponds to the difference of their section ID.

$$dists(p_i, q_i) = \left| SecId(p_j) - SecId(q_j) \right| + 1 \tag{1}$$

Definition(1) is used to measure the points' scatter in the second projections. In the dimension *i* before second projection, assume that the points *p* and *q* are in the same section. After applying the second projection from the dimension *i* to *j*, assume that the points *p* and *q* are located in different sections with different section IDs. So we can compare the distance of two points by the subtraction between $SecId(p_j)$ and $SecId(q_j)$ as in (1). The $dists(p_i, q_i)$ is defined as the absolute difference value between the two points' sections, plus 1 in order to avoid the computational complexity of 0. In k-NS algorithm, *dists* supplies the effective factor to evaluate the scatter of the points in the second projected dimensions.

The definition of outliers in k-NS is regarded as a statistic weight value, which is decided by its related calculated results in all projected dimensions.

**DEFINITION 2** (Outlier in k-NS)
The $x_{ns}$ of a given point x$\in Section$ in the database D$\subset \mathbb{R}^m$ is defined as follows:

$$x_{ns} = \{x, x' \in D \,\middle|\, \forall x \in D, x, x' \in Section_i, \forall p \in Section_i$$
$$\sum_{i=1}^{m} d_i(x) << \sum_{i=1}^{m} \overline{d_i} \cup \sum_{i=1}^{m} \sum_{j=1, j \neq i}^{m} dists(x_j', p_j) \leq \sum_{i=1}^{m} \sum_{j=1, j \neq i}^{m} dists(x_j, p_j)\} \tag{2}$$

*x*, *x*\` and *p* are the points in the same section in the dimension *i*. The point *p* is any of the neighbor points by the measure of *dists* after applying the second projection. $x_{ns}$ is a statistical result for summarizing all the values of *x* in two dimensional-projections, which means $x_{ns}$ of $x$ can be used as a final result to detect outliers.

By the k-NS definition, outliers satisfying either of the following two conditions are to be detected: first, outliers that can be detected in the first projection; second, outliers that still can be detected by *dists*-based k-NS in the second projection even if the point does not appear abnormal in first projection.

Although the $x_{ns}$ in (1) can reflect the outlier result, it is difficult to be calculated for each point. Therefore, the general statistic information for each point is defined in (3):

**DEFINITION 3** (General Statistical Information of Point) Set $sdr_{Proj_i}(p_{i,k})$ is the calculated value of $p_k$ in the first projected dimension and $sdr_{Proj_{i\to j}}(p_{i,k})$ is the calculated value of $p_k$ after second projection from the dimension $i$ to the dimension $j$. $\omega_1$ and $\omega_2$ are the weight parameters for these two values. Then the statistical information value of $p_i$ is expressed as follows:

$$SI_{ns}(p_k)=\left\{\omega_1\sum_{i=1}^{m}sdr_{Proj_i}(p_{i,k})+\omega_2\sum_{i=1}^{m}\sum_{j=1,j\neq j}^{m}sdr_{Proj_{i\to j}}(p_{j,k})\right\} \quad (3)$$

*sdr* is used to calculate the density ratio of point in two dimensional-projections. The detail calculation method is introduced in (4) and (5). *SI* (Statistic Information) is the point's final score by which all the points are evaluated. The outlier's SI value is obviously different from normal point's. For the different dataset, adjust the weight values may bring the better result.

*1) Section Density Ratio Calculated in first Projected Dimension*

Outliers always appear more sparsely than most normal points if they can be detected in projected dimensions. Therefore, the section density of outliers is lower than the average section density in that dimension.

In our proposal, *sdr*(Section Density Ratio) is cited for calculation. The *sdr* of a point not only reflect the sparsity compared with others in that dimension, but also keep this value independent between different dimensions.

**DEFINITION 4** (Section Density Ratio)
Set point $p_{i,j}\in Section_{i,\gamma}$ in the dimension $i$, where $j$ is the point ID and $\gamma$ is the section ID. $d_{i,\gamma}$ is the section density of point $p_{i,j}$ in dimension $i$, and $\overline{d_i}$ is the average section density in dimension $i$. The $p_{i,j}$'s *sdr* is denoted by *sdr* of $Section_{i,\gamma}$, which is defined as follows:

$$sdr_{Proj_i}(p_{i,j})\Leftarrow sdr_{Proj_i}(Section_{i,\gamma})=\left(\frac{d_{i,\gamma}}{\overline{d_i}}\right)^2 \quad (4)$$

One point to be noticed is that one *sdr(section)* does not only correspond to one point, but it is shared by all the points in the same section. Hence, the section $sdr_{Proj_i}(Section_{i,\gamma})$ is assigned to the point $sdr_{Proj_i}(p_{i,j})$. Totally, $m$-times $sdr_{Proj_i}$ are obtained from all dimensions for each point.

**Lemma 1.** Given a data set DB and point $p$ of DB in a section of the dimension $i$, $Card(Section_i)=scn$,

$d(p)=Count(Section(p))$ and $d_i=\frac{1}{scn}\sum_{k=1}^{scn}Count(Section_{i,k})$,

if $p$ is an outlier, then $\frac{d_i(p)}{\overline{d_i}}<1$.

Where $Card(Section_i)$ is the number of sections in dimension $i$. $Section(p)$ refers to the section the point $p$ is in. $Count(Section(p))$ is the number of points in $Section(p)$.

**Proof.** First, we set the outlier $p$ is not in a normal's cluster in projected dimensions, otherwise Definition(4) is to be applied.

$\forall q$, $Count(Section(p))\leq Count(Section(q))$

$$d_i(p)=Count(Section(p))\leq\frac{1}{n}\sum_{j=1}^{n}Count(Section(q))$$

Since the section density $d$ of outlier $p$ is less than most of points' according to outlier's definition,

$$d_i(p)\prec\frac{1}{n}\sum_{j=1}^{n}Count(Section(q_j))$$
$$=\frac{1}{n}\sum_{k=1}^{c}Count(Section_k(\sum q))$$
$$=\frac{1}{c}\sum_{k=1}^{c}Count(Section_k)=\overline{d_i}$$

So $\frac{d_i(p)}{\overline{d_i}}<1$

*2) k-Nearest Section Calculated in Second Projected Dimension*

If outliers do not appear clearly in the low dimensions, they cannot be detected by the first step since they are hidden among the normal points and have similar distance or density with others. Nevertheless, these points still can be detected in the second projected dimensions. This step aims to find outliers from normal points by projecting these points into different dimensions. The section distance measurement describes the sparsity of points to check them in second projected dimensions. Basing on the section distance concept and referring to the k-Nearest Neighbor concept[10], we can get the *sdr* of the nearest sections of the point in the projected dimensions.

**DEFINITION 5** (Nearest Sections in Projected Dimension) In the second dimension projection, the dimension is projected from $i$ to $j$, Set $p_j,\ p_f,\ q\in section_i$, the nearest section neighbor $N_{kn}(p)$ of the point $p$ is defined as $N_{kn}(p)=\{q\in section_i \mid dists(p,q)\leq k-dists(p)\}$. The point $q$ is one of k-nearest neighbor points. $Count(Section_{i,\gamma})=s$.

$|N_{kn}|$ is the number of $p$'s neighbors. Then $sdr_{Proj_{i\to j}}$ of point $p_{i,k}$ is defined as follows:

$$sdr_{Proj_{i\to j}}(p_{i,k}) \Leftarrow sdr_{Proj_{i\to j}}(Section_{i,\gamma}) = \frac{\frac{1}{|N_{kn}|}\sum_{q\in N_{kn}} dists(p_{j,k},q))^2}{\frac{1}{s}\sum_{f=1}^{s}\left[\frac{1}{|N_{kn}|}\sum_{q\in N_{kn}} dists(p_{j,f},q))^2\right]} \qquad (5)$$

We calculate the $p_k$'s *dists* with k-nearest neighbor points, and then get the ratio value against the average value of points' in the same section. While the point is projected to another dimension, a single $sdr_{Proj_{i\to j}}$ value is calculated each time of the projection. Totally, $m\times(m-1)$ -times $sdr_{Proj_{i\to j}}$ values are obtained from all the projected dimensions for each point.

**Lemma 2.** Given a dataset DB and point $o,p,q \in Section_i$ in a dimension *i*. Set a normal points' cluster C, normal points $p,q\in C$ , after the second projection, points *p*, *q*, *o* are projected to dimension *j*. *p* is *o*'s the k^th nearest neighbor, *q* is *p*'s the k^th nearest neighbor.

If o is an outlier, then $dists(o,p)\geq dists(p,q)$

**Proof.** For normal points, they belong to a cluster in all dimensions. Therefore, $p,q\in C$ in dimension *j*. *o* is an outlier, so $o\notin C$ .

If *q* is in the *o*'s k neighbors,

Then If p, q are on the *o*'s same side $dists(o,p)\geq dist(p,q)$

If p, q are on the *o*'s both sides, $\because p,q\in C \quad \therefore o\in C$

$\Rightarrow \quad o\in C \cap o\notin C \rightarrow \varnothing$ , it is the contradiction!

If *q* is not in the *o*'s k neighbors, *q* is on the other side of *p*

If $dists(p,q) > dists(o,p) \Rightarrow o\in C$

$\Rightarrow o\in C \cap o\notin C \rightarrow \varnothing$ , it is the contradiction!

Then, $dists(o,p) > dists(p,q)$

#### 3) *Statistical Information Values for Each Point*

Through the above two steps calculation, each point gets *m*-times $sdr_{Proj_i}$ values at first projection and gets $m\times(m-1)$ -times $sdr_{Proj_{i\to j}}$ values in second projection.

The suitable weights for *SI* in (3) are considered in order to give the sharp boundary to compare points. By evaluating different weighting values and their performance, we choose simple and clear values. Here, we get the reciprocal value of average $sdr_{Proj_i}$ and $sdr_{Proj_{i\to j}}$ , so we set weight $\omega_1 = \frac{1}{m}$ and $\omega_2 = \frac{1}{m(m-1)}$. The outliers have obviously larger *SI* than that of the normal points'.

**DEFINITION 6** (Statistic Information of point)

$$SI_{ns}(p_k) = \frac{2m}{\sum_{i=1}^{m}\left[sdr_{Proj_i}(p_{i,k}) + \frac{1}{m-1}\times\sum_{j=1}^{m-1} sdr_{Proj_{i\to j}}(p_{i,k})\right]} \qquad (6)$$

Equation(6) sums up *sdr* values in all projected-dimensions. In low dimension, the *SI* value for normal points should be close to 1, and outlier's *SI* value should be obviously larger than 1. However, it is not true in high dimensional space. Normal points' *SI* are getting close to outlier's *SI*. Nevertheless, the outlier's *SI* is still obviously higher than normal points'. Therefore, outliers can be detected just by finding points with top largest *SI* values.

### E. *Algorithm*

Now, we focus on how to implement the k-NS method in R language. How to get *PointInfo* and *SectionInfo* effectively in different sections and dimensions is a key issue that needs to be considered in detail. The proposed algorithm is shown in Table 2 with pseudo-R code. Here, the dataset has *n* points in *m* dimensional space. The range of data is divided into *scn* sections in each dimension.

Table 2. k-NS Algorithm

| **Algorithm: k-Nearest Section** |
|---|
| **Input:** k, data[n,m], scn |
| Begin |
| Initialize(PointInfo[n, m], SectionInfo[scn, m]) |
| For i=1 to m |
| $d_i$=n/length(SectionInfo[ SectionInfo[i,]!=0,i]) |
| For j=1 to n |
| Get $sdr_{Proj_i}(Section_{i,\gamma})$ with section density ratio in (4) |
| $sdr_{Proj_i}(Section_{i,\gamma})$ denote $sdr_{Proj_i}(p_{i,j})$ (PointInfo[i,j]=$\gamma$) |
| End n |
| End m |
| For c=1 to 10 |
| resort dimension i in random order |
| For i=1 to m |
| For j=1 to scn |
| PtNum <- SecInfo[j,i] |
| If(PtNum ==0) next |
| Ptid <- which(PtInfo[,i]==j) |
| If(PtNum <$\frac{3}{2}\times k$) $sdr_{Proj_{i\to j}}$ =1 |
| else For each(p in Ptid]) |
| { |
| if ( i<m) i=i+1 |
| else i=1 |
| Get dists($P_{Ptid,\ i}$) with Definition(1) |
| Get $sdr_{Proj_{i\to j}}$ with Definition(5) |
| } |
| End j |
| End i |
| End c |
| Get SI value with Definition(6) for each point |
| **Output**: Outliers with Point ID (SI(p) >> $\overline{SI}$ or top SI score) |

Three points need to be clarified in this algorithm.

The first point is how to decide the average section-density $\overline{d_i}$ in each dimension. $\overline{d_i}$ value is obtained by the definition of the average section density $\frac{n}{scn}$. It means $\overline{d_i}$ is same in each dimension. However, we consider the special case that most points are in several sections and no point is in other sections. In this case, $\overline{d_i}$ becomes very low and even close to the outlier's section density. Therefore, we only count sections with points. Subsequently, $\overline{d_i}$ are varied in different dimensions. Hence, the ratio of the section density against $d_i$ in Definition (4) can measure the sparsity of points in different sections of a dimension.

The second point is the number of points in one section. There are three different cases.

- Case 1: no point in the section. In this case, the algorithm just passes this section and goes to the next section.
- Case 2: many points in the section. In this case, the nearest sections method is used directly to detect points.
- Case 3: only a few points in the section. In this case, the point distribution is difficult to be judged just by these several points. In addition, the section density ratio in the step 1 must be very low. Therefore, these points are to be already detected by the previous step. Here, we pass this section too.

The threshold value to separate the case 2 and case 3 is related to the *k*. *k* should not be large because *k* is less than $\overline{d_i}$ in the step 2. Through experiments with values from 4 to 20 to find the suitable value for *k* and the threshold of the number of points in one section, we have found that the threshold value can be defined as $\frac{3}{2} \times k$ as the best solution which could be used in most of the situations.

### F. *Complexity Analysis*

The three-step procedure is considered separately to state the complexity of the k-NS algorithm. In the first step, it calculates the section density in each projected dimension. The time complexity is $O(m \times n)$. In the second step, k-nearest sections density is calculated between projected dimensions. The time complexity is $O(scn \times (m-1) \times m)$. It is noticed that all points in the section are used, so the time complexity expression is changed to $O(n \times (m-1) \times m)$. In the last step, summing up all weight values for each point, the time complexity is O(n). Hence, the total complexity time is

$$T(n) = O_1 + O_2 + O_3$$
$$= O(m \times n) + O(n \times (m-1) \times m) + O(n)$$
$$\approx O(n \times m^2)$$

The k-NS takes more processing time on calculating in the loop of dimension projections and finding the related point's section in each dimension.

The space complexity of k-NS is

$$S(n) = O(2 \times scn \times m + 3 \times m \times n + m + n)$$
$$\approx O(3 \times m \times n)$$

We need to record the necessary information and intermediate result for the point and the section. The temporary room needed in the procedure is just a little.

### G. *Distinction between Outliers and Noisy Points*

The concept of outlier and noisy point has been proposed for more than ten years. According to that, outlier is regarded as abnormal data, which is generated by a different mechanism and contains valuable information, and noisy data are regarded as a side product of clustering points, which have no useful information but affect the correct result greatly.

In the data space, outliers are the points that are farther from others by some measures, while the noisy points always appear around the outliers. Since the noisy points are also far away from the normal points, in low dimensional space, it is difficult to make a distinct boundary between outliers and noisy points. Based on this frustrated observation, some researchers even consider that noisy point is as a kind of outliers. There is no difference in detecting abnormal data by any methods. Hence, it is a meaningful issue to make them different between outliers and noisy points not only in concept but also in detection measures.

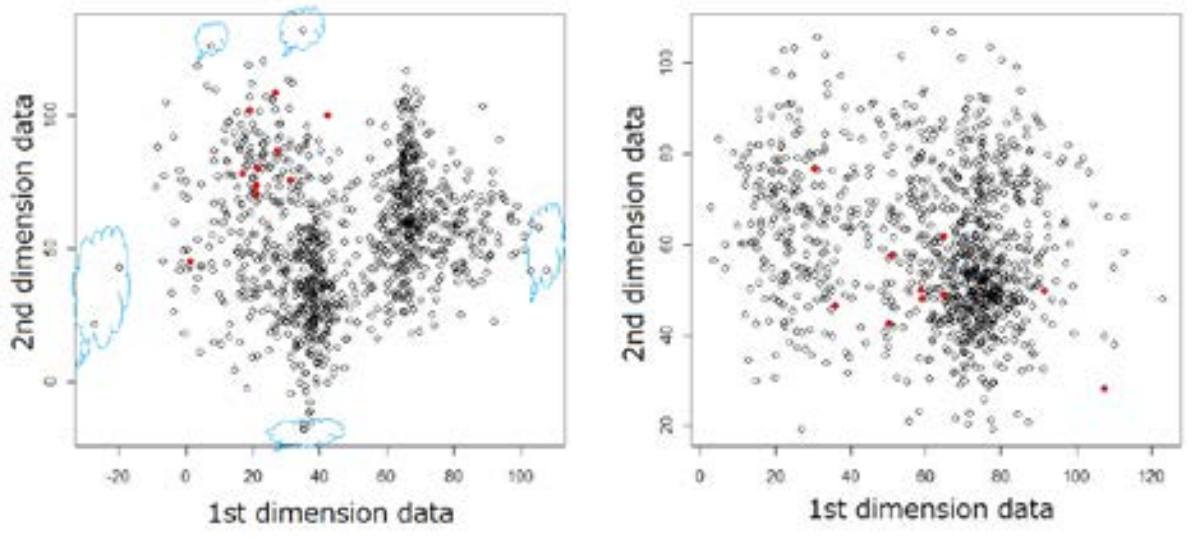

(a) (b)

Figure 3. Noisy Data of Dataset 5 Projected to Two-Dimensional Space and Dataset 3 Projected to Two-Dimensional Space

In this paper, we try to explain the distinction between outliers and noisy points in two aspects. The first is that there are different data generation processes. Outliers are generated by a different distribution from normal points. Noisy points have the same distribution with normal points. The second is that abnormal states are different in dimensional space. Outliers appear abnormal in most of the dimensions. Noisy points only appear abnormally in several dimensions and appear normal in other dimensions. From the whole dimensions' view, these noisy data also conform to the same distribution of normal data. The outlier may appear in the same way in low dimensional space, but they conform to different distribution mechanism from normal points'. Therefore, it shows the difference between outliers and noisy points in some projected-dimensional spaces. An example of noisy data is shown in Fig. 3(a). The data is retrieved from Dataset 8 as introduced in section 4, which contains 1000 points in 10000 dimensions. The outliers are placed in the middle region and can be found differently from normal points. The noisy points are labeled with a cloud symbol that is so different in this projected two-dimensional space. Another example is shown in Fig. 3(b). The outliers are not always obvious in low projected dimensional space, while noisy points that are distributed on the marginal area of both dimensions are likely abnormal points.

## IV. EVALUATION

We have implemented our algorithm and applied it to several high dimensional datasets, and then have made the comparison between k-NS, LOF and LOCI. In order to compare these algorithms under fair conditions, we performed them with R language, on a Mac Book Pro with 2.53GHz Intel core 2 CPU and 4G memory.

### A. *Synthetic Datasets*

A critical issue of evaluating outlier detection algorithms is that there are no benchmark datasets available in a real world to satisfy the explicit division between outliers and normal points. The points that are found as outliers in some

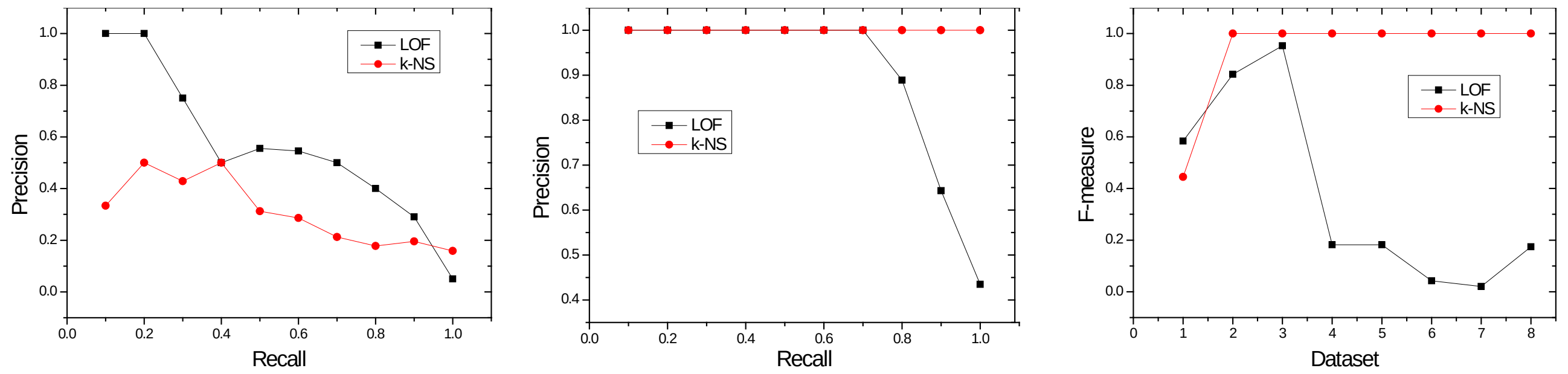

(a) Precision-Recall in Dataset 1with 10 Dimensions (b) Precision-Recall in Dataset 2 with100 Dimensions (c) F-measure of Dataset 1-8

Figure 4. Effectiveness Comparison between LOF and k-NS in Eight Datasets from dimension 10 to 10000. Precision-Recall in (a) and (b); F-measure in (c)

Table 3. Experiment Dataset

| Dataset No. | Dimension number | Points number | Normal points (Normal distribution) | Outliers Random |
|---|---|---|---|---|
| 1 | 10 | 500 | μ(20-80), σ(10-20) | (20-100) |
| 2 | 100 | 500 | μ(20-80), σ(10-20) | (20-100) |
| 3 | 100 | 1000 | μ(20-80), σ(10-20) | (20-100) |
| 4 | 500 | 500 | μ(20-80), σ(10-20) | (20-100) |
| 5 | 500 | 1000 | μ(20-80), σ(10-20) | (20-100) |
| 6 | 1000 | 500 | μ(20-80), σ(10-20) | (20-100) |
| 7 | 1000 | 1000 | μ(20-80), σ(10-20) | (20-100) |
| 8 | 10000 | 1000 | μ(20-80), σ(10-20) | (20-100) |

real dataset are impossible to provide a reasonable explanation why these points are picked out as outliers. On the other hand, what we have learned from the statistical knowledge is helpful to generate the artificial dataset: if some points with some distributions are apparently different from those of normal points, these points can be regarded as outliers. Hence, we generate the synthetic data based on this assumption.

We generate the eight synthetic datasets with points of 500-1000 and dimensions of 10-10000. The normal points conform to the normal distributions while outliers conform to the random distributions in a fixed region. Normal points are distributed in five clusters with random $\mu$ and $\sigma$, and 10 outliers are distributed randomly in the middle of normal points' range. The more details about the parameters in each dataset are shown in Table 3.

The experiment datasets are generated by the rules that outliers' range should be within the range of the normal points in any dimensions. Therefore, outliers cannot be found in low dimensional space. The data distribution example is shown in Fig. 3(b) where the Dataset 3 is projected to two-dimensional space with outliers labeled with red color. It is clearly shown that the outliers are within the range of normal points and appear no difference with the normal points in this two-dimensional space. Noisy points that are placed on the margin of distributed area are more likely regarded as abnormal points. Hence, outliers and normal data cannot be separated just by the straight observation of the different distributions.

## B. *Effectiveness*

First, we conduct the two-dimensional experiment using the dataset in Fig. 2. The result show that three algorithms perform well. Our proposed algorithm can run on low dimensional dataset.

Next, our proposed algorithm is evaluated thoroughly by a series of experiments and compared it with LOF. LOCI is excluded for comparison because it performs poor in every dataset. In order to measure the performance of these algorithms with precision and recall, the 10 outliers are reprieved one by one. In the evaluation of all eight dataset experiments, we obtained 10 precisions and the 10 recalls respectively in every dataset, and obtain the 10 F-measures. We pick up the highest F-measure from each dataset for demonstrating the experiment performance by LOF and k-NS.

At the beginning, we need to set all the appropriate parameters in the eight experimental datasets. The parameters are the best ones for the prepared datasets, and they are changed according to the data size and the number of dimensions. The parameter Knn of LOF is set around 10 in all the experiments since the dataset size is only 500 or 1000 points. This is a reasonable ratio of neighbor points against the whole dataset size. For our algorithm, the parameters of $\bar{d}$ and *scn* are inverse each other. The product of $\bar{d}$ and *scn* is equal to *n*. We set *scn* a little larger than $\bar{d}$ , because these combinations of parameters have shown the better experiment results.

The 10-dimensional experiment result is shown in Fig. 4(a). LOF performs best in this 10-dimensional experiment. Especially, LOF can detect two outliers with very high precision. Nevertheless, the precision of LOF falls down sharply with the increasing recall from 20% to 40%. At last, the result of precision is worse than k-NS in detecting all outliers correctly. As a whole, the performance of k-NS is below LOF. The reason that performance is poor in both algorithms is that the outliers are placed in the center of normal data in our datasets, which prevents these outliers to be found in low dimensional space. Therefore, it is difficult to find exact outliers in 10-dimensional space.

When the number of dimension increases to 100, the precision and recall in the $2^{nd}$ dataset clearly show the

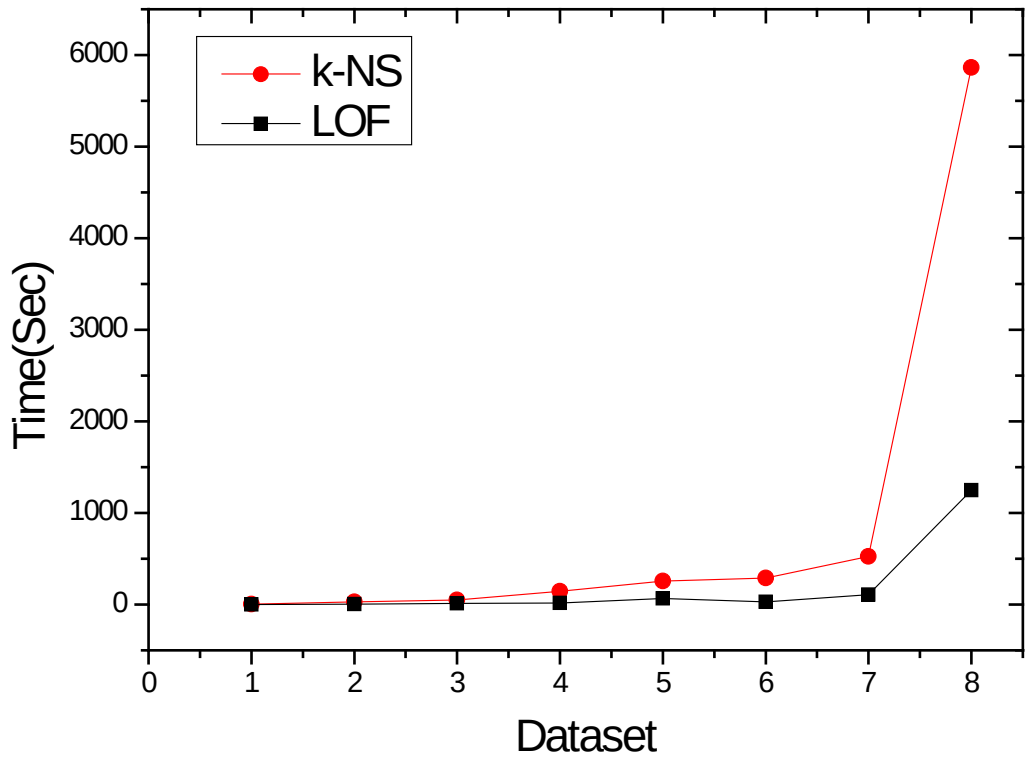

Figure 5. Running Time

| No. | LOF | | K-NS | |
|---|---|---|---|---|
| | SI | Pt ID | SI | Pt ID |
| 1 | 1.432643 | 315 | 0.06458754 | 189 |
| 2 | 1.394097 | 311 | 0.06294790 | 221 |
| 3 | 1.338282 | 140 | 0.06181901 | 314 |
| 4 | 1.279471 | 163 | 0.06152080 | 132 |
| 5 | 1.230594 | 115 | 0.05985524 | 237 |
| 6 | 1.196492 | 110 | 0.05924606 | 34 |
| 7 | 1.193774 | 8 | 0.05894665 | 190 |
| 8 | 1.175922 | 317 | 0.05754411 | 50 |
| 9 | 1.172092 | 105 | 0.05748593 | 309 |
| 10 | 1.156304 | 245 | 0.05737442 | 242 |
| 11 | 1.155840 | 312 | 0.05633002 | 180 |
| 12 | 1.146046 | 313 | 0.05614289 | 78 |
| 13 | 1.140580 | 263 | 0.05590464 | 111 |
| 14 | 1.134468 | 314 | 0.05566309 | 48 |
| 15 | 1.132395 | 112 | 0.05548088 | 131 |
| 16 | 1.124824 | 316 | 0.05533588 | 318 |
| 17 | 1.118983 | 224 | 0.05514495 | 126 |
| 18 | 1.118789 | 137 | 0.05485238 | 29 |
| 19 | 1.115959 | 180 | 0.05483668 | 192 |
| 20 | 1.109533 | 63 | 0.05476878 | 92 |

a

| Pt | LOF | | k-NS | | Mixed | |
|---|---|---|---|---|---|---|
| | Pt | Recall | Pt | Recall | Pt | Recall |
| 10 | 3 | 30% | 2 | 20% | 5 | 50% |
| 20 | 7 | 70% | 3 | 30% | 9 | 90% |

b

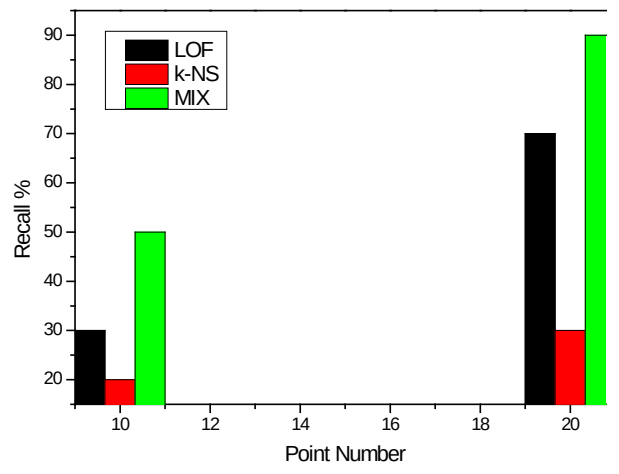

c

Figure 6. Top 20 Points Detected in Arcane Data

effectiveness of these algorithms. Different from the first dataset, k-NS achieves 100% precision with any recall all the time. LOF obviously reduces the precision from 100% to 43.48% with the increasing recall from 70% to 100%, as shown in Fig. 4(b). In fact, k-NS keeps the perfect result in 100 dimensions, while LOF performs much poorer in terms of the precision and the recall.

The experiments of the datasets from 1 to 8 are shown in Fig. 4(c). LOF needs to pick the largest F-measure for each dataset, while k-NS only needs to pick the largest F-measure for the first dataset. In addition, F-measures of k-NS are always 1 on the datasets 2 to 8.

The experiments show that k-NS performs perfectly in find inner outliers in high dimensional space. LOF is suffered the curse of high dimension greatly. We find that the precision become better when the dataset size is increased; but it does not for LOF.

### C. *Efficiencies*

We also compare these algorithms in running-time. In R language, the running time includes user time, system time and total time. So we only use the user time to compare them.

As shown in Fig. 5, LOF is faster in all experiments. The two algorithms take more time when the number of dimensions or the data size increase. The reason is that there is no dimension-loop calculation for LOF because it only processes the distance between a point and its neighbors. However, our proposed algorithm calculates the values in all the first projected dimensions and all the second projected-dimensions.

### D. *Performance on Real World Data*

In this subsection, we compare these algorithms with a real-world dataset publicly available at the UCI machine-learning repository[24]. We use Arcene dataset that is provided by ARCECE group.

The task of the group is to distinguish cancer versus normal patterns from mass-spectrometric data. This is a two-class classification problem with continuous input variables. This dataset is one of five datasets from the NIPS 2003 feature selection challenge.

The original dataset includes total 900 instances with 10000 attributes. The datasets have training dataset, validating dataset and test dataset. Each sub-dataset is labeled with positive and negative except for test dataset. For 700 instances in test dataset, we only know 310 instances are positive and 390 instances are negative. The best_SVM_result is available at [25]. 308 instances are labeled with positive, and 392 instances are labeled with negative. We use this SVM result for evaluating LOF and our proposal, where we create a dataset by adding randomly selected 10 negative instances to the retrieved 308 positive instances by SVM. The first evaluation uses this dataset with total 318 instances. The second evaluation uses the retrieved 392 negative instances, and we apply two algorithms to detect outlier from them.

The result of the first experiment is shown in Fig. 6. The 20 top points are chosen in both algorithms. SI is the score for point, and the Pt ID is point ID. The points with *Pt ID* larger than 308 are true outliers. Three outliers and two outliers are detected in LOF and k-NS in the top ten points. Totally, five outliers are detected by mixed result combined with both algorithms. In the top twenty points, seven outliers and three outliers are detected by LOF and k-NS. Totally, nine outliers are detected with mixed result. In both results, the LOF is better than k-NS. However, the k-NS can help to increase the detection accuracy from 30% to 50% in 10 points, 70% to 90% in 20 points. In another word, the k-NS supply a reasonable alternative solution to increase the precision results. As a contrast, we also give the LOCI result, which output point ID (8, 20, 48, 95, 153, 189, 193, 242, 307, 311, 315, 317). Its recall is 30%, the same as k-NS. However, all the outliers detected in LOCI are also detected by LOF.

Table 4. Top 5 Points detected in Arcane Data

| No. | LOF | | K-NS | | LOCI |
|---|---|---|---|---|---|
| | SI | Pt ID | SI | Pt ID | Pt ID |
| 1 | 1.477314 | 182 | 0.05711392 | 29 | 153 |
| 2 | 1.364228 | 175 | 0.05643270 | 318 | 175 |
| 3 | 1.347864 | 29 | 0.05430054 | 182 | 195 |
| 4 | 1.321561 | 153 | 0.05281425 | 215 | |
| 5 | 1.317432 | 360 | 0.05047134 | 234 | |

In the second experiment, there are two positive points miss-clarified by SVM. Therefore, finding these two points is the task of this experiment. As seen in Table 4 showing the results, point IDs of 29 and 182 are the most probably outliers by the intersection of LOF and k-NS results. It is noted that the both points appear in the top three detected points of the both results. If we consider the LOCI result, the intersections point ID is 153 and 175, which is entirely different from k-NS. Nevertheless, contrast with former results, the first conclusion seems more reasonable.

## V. CONCLUSTION

In this paper, we introduce a new definition of inner outlier, and then present a novel method, called k-NS, designed to detect such inner outliers with the top largest score in a high dimensional dataset. The algorithm is based on a statistical method with three steps. (i) Calculate the section density ratio of each point in each dimension after first projection. (ii) Compute the nearest sections density ratio of each point in all projected dimensions after second projection. (iii) Summarize all *sdr* values of each point and denoted as a weight value (SI), then compare SI with those of the other points. Each point gets totally $m+m\times(m-1)$ values to be compared. Experimental results on synthetic datasets with dimension from 10 to 10000 have shown that our proposed k-NS algorithm has the following advantages:

- Immune to the curse of high dimensions,
- Adapt to various outlier distributions,
- Show outstanding performance on detecting inner outliers in high dimensional data space.

The difference between outliers and noisy data is also discussed in this paper. This issue is difficult in low dimensional space. In our experiments, the noisy data and outlier are found differently by comparing the distribution in projected dimension and whole dimensions, and the noisy data seem to more abnormal than outliers in some projected dimensional spaces in our cases.

As the ongoing and future work, we continue to improve the algorithm by finding the best relationship for two-step *sdr*. Besides performing the dataset with the high dimensions, the dataset with large-scale data size or increment updates instead of computing it over the entire dataset to the outlier detection need to be conducted. Another issue is the expensive cost of the processing time in high dimensional space. Any solution to reduce the processing time needs to be investigated. One of the approaches may be the use of the parallel processing.

## REFERENCES

[1] Markus M.Breunig, Hans-Peter Kriegel, Raymond T.Ng, Jorg Sander. LOF: Indetify density-based local outliers. Proceedings of the 2000 ACM SIGMOD international conference on Management of data.

[2] Charu C.Aggarwal, Philip S.Yu. Outlier detection for high dimensional data. Proceedings of the 2001 ACM SIGMOD international conference on Management of data.

[3] Spiros Papadimitriou, Hiroyuki Kitagawa, Phillip B.Gibbons. LOCI: fast outlier detection using the local correlation integral. IEEE 19th International conference on data engineering 2003.

[4] Hans-peter Kriegel, Matthias Schubert, Arthur Zimek. Angle-based outlier detection in high dimensional data. The 14th ACM SIGKDD international conference conference on Knowledge discovery and data mining. 2008.

[5] Christian Bohm, Christos Faloutsos, etc. Robust information theoretic clustering. The 12th ACM SIGKDD international conference conference on Knowledge discovery and data mining. 2006.

[6] Zhana, Wataru Kameyama. A Proposal for Outlier Detection in High Dimensional Space. The 73rd National Convention of Information Processing Society of Japan, 2011.

[7] D. Hawkins. Identification of Outliers. Chapman and Hall, London, 1980.

[8] Fabrizio Angiulli and Clara Pizzuti. Outlier mining in large high-dimensional data sets. IEEE Transactions on Knowledge and Data Engineering (TKDE), 17(2):203-215, February 2005.

[9] Christian Bohm, Katrin Haegler. CoCo: coding cost for parameter-free outlier detection. In Proceedings of the 15th ACM SIGKDD international conference conference on Knowledge discovery and data mining. 2009.

[10] Alexandar Hinnerburg, Charu C. aggarwal, Daniel A. Keim. What is the nearest neighbor in high dimensional space? Proceedings of the 26th VLDB Conference, 2000.

[11] Dantong Yu,etc. FindOut: finding outliers in very large datasets. Knowledge and Information System (2002) 4:387-412.

[12] Christian Bohm, Christos Faloutsos, etc. Outlier-robust clustering using independent components. Proceedings of the 2008 ACM SIGMOD international conference on Management of data.

[13] De Vries, T., Chawla, S., Houle, M.E., Finding Local Anomalies in Very High Dimensional Space, 2010 IEEE 10th International Conference on Data Mining(ICDM), pp.128-137, 13-17 Dec. 2010.

[14] Anny Lai-mei Chiu and Ada Wai-chee Fu, Enhancements on Local Outlier Detection. Proceedings of the Seventh International Database Engineering and Applications Symposium (IDEAS'03)

[15] Aaron Ceglar, John F.Roddick and David M.W.Powers. CURIO: A fast outlier and outlier cluster detection algorithm for larger datasets. AIDM '07 Proceedings of the 2nd international workshop on Integrating artificial intelligence and data mining. Australia, 2007.

[16] Feng chen, Chang-Tien Lu, Arnold P. Boedihardjo. GLS-SOD: a generalized local statistical approach for spatial outlier detection. . In Proceedings of the 16th ACM SIGKDD international conference on Knowledge discovery and data mining. 2010.

[17] Michal Valko, Branislav Kveton, etc. 2011. Conditional Anomaly Detection with Soft Harmonic Functions. In Proceedings of the IEEE 11th International Conference on Data Mining (ICDM '11).

[18] Ashok K. Nag, Amit Mitra, etc. Multiple outelier detection in multivariate data using self-organizing maps title. Computational statistics. 2005.20:245-264.

[19] Teuvo kohonen. The self-organizing map. Proceedings of the IEEE, Vol.78, No.9, September, 1990.

[20] Naoki Abe, Bianca Zadrozny, John Langford. Outlier Detection by Active Learning. Proceedings of the 12th ACM SIGKDD international conference. 2006.

[21] Ji Zhang, etc. Detecting projected outliers in high dimensional data streams. In Proceedings of the 20th International Conference on Database and Expert Systems Applications (DEXA '09).

[22] Alexander Hinneburg, Daniel A. Keim. Optimal grid-clustering:Towards breaking the curse of dimensionality in high dimensional clustering. The 25th VLDB conference 1999.

[23] Amol Ghoting, etc. Fast Mining of Distance-Based Outliers in High-Dimensional Datasets. Data Mining and Knowledge Discovery. Vol.16:349-364, 2008.

[24] http://archive.ics.uci.edu/ml/datasets/Arcene(visited on May6th,2012).

[25] http://clopinet.com/isabelle/Projects/NIPS2003/analysis.html#svm-resu (visited on May 6th,2012).